# A Comparative Analysis of Techniques and Algorithms for Recognising Sign Language

Rupesh Kumar
Department of CSE
Galgotias College of Engineering and Technology, AKTU
Greater Noida, India
rskumar0402@gmail.com

Ashutosh Bajpai
Department of CSE
Galgotias College of Engineering and Technology, AKTU
Greater Noida, India
ashubajpai161@gmail.com

Ayush Sinha
Department of CSE
Galgotias College of Engineering and Technology, India
Greater Noida, India
ayushsinha132001@gmail.com

S.K Singh
Professor Department of CSE
Galgotias College of Engineering and Technology, AKTU
Greater Noida, India
Drshashikant.singh@galgotiacollege.edu

*Abstract*—Sign language is a visual language that enhances communication between people and is frequently used as the primary form of communication by people with hearing loss. Even so, not many people with hearing loss use sign language, and they frequently experience social isolation. Therefore, it is necessary to create human-computer interface systems that can offer hearing-impaired people a social platform. Most commercial sign language translation systems now on the market are sensor-based, pricey, and challenging to use. Although vision-based systems are desperately needed, they must first overcome several challenges.

Earlier continuous sign language recognition techniques used hidden Markov models, which have a *limited* ability to include temporal information. To get over these restrictions, several machine learning approaches are being applied to transform hand and sign language motions into spoken or written language. In this study, we compare various deep learning techniques for recognising sign language. Our survey aims to provide a comprehensive overview of the most recent approaches and challenges in this field.

## I. INTRODUCTION

The study of sign language is a significant and intriguing field of study that has effects on both the deaf population and society at large. Researchers are interested in studying sign language for several reasons. Firstly, a better understanding of sign language can lead to better communication between the deaf community and the general public. Secondly, researching sign languages can provide insight into the nature of language itself and the interaction between language and the brain. Lastly, a better understanding of sign language can lead to the development of more effective sign language instruction techniques for both deaf and hearing people.

Sign languages are used in many countries, and each has its own distinctive vocabulary and grammar, such as British Sign Language (BSL) in the UK and American Sign Language (ASL) in the US. However, they all share commonalities in the way that meaning is expressed through hand and body gestures. Sign languages have their own set of phonological, morphological, and syntactic principles, and they are not just visual representations of spoken languages. They are independent languages with their own syntax, vocabulary, and grammar, and their many gestures, facial expressions, and body positions convey different meanings.

The significance of sign language is highlighted by the World Health Organization's estimate that there are more than 70 million deaf or hearing-impaired persons worldwide. Many nations, including the United States, Canada, and Australia, recognize sign language as an official language, underscoring the importance of recognizing, researching, and preserving sign language as a unique and independent language. The use of sign language can help reduce linguistic barriers, promote inclusion, and improve accessibility for deaf and hearing-impaired people in various contexts.

In conclusion, sign language is a vital tool for the global deaf community. Understanding and acknowledging sign language as a unique and independent language is crucial to promoting accessibility and inclusivity. Sign language research can shed light on language and the brain, and effective sign language instruction techniques can benefit both deaf and hearing people.

This study attempts to analyse recent developments in deep learning-based sign language recognition (SLR), including methods and prospective applications. The authors reviewed deep learning architectures and algorithms for SLR and evaluated their performance, highlighting its importance for the deaf community. The study provides valuable information about the latest developments in SLR and its potential applications, serving as a guide for researchers and practitioners in the field.

## II. RELATED WORK

In this section, we review relevant research papers in sign language recognition techniques, classifying them into glove-based and vision-based approaches. The glove-based techniques involve wearable devices like data gloves to capture hand movements and gestures, while vision-based techniques utilize computer vision algorithms to analyse visual cues. By examining these studies, we aim to provide a comprehensive overview of the existing techniques and their implications for advancing sign language recognition technology.

*A. Glove-based approach*

In their paper, Khomami et al. [21] developed a wearable hardware using surface electromyography (sEMG) and Inertial Measurement Unit (IMU) sensors for Persian Sign Language (PSL) recognition. The system's ability to accurately capture signs was increased by fusing these two sensors. By extracting and classifying the 25 highest-ranked features using the KNN classifier, they achieved an average accuracy of 96.13%. Their affordable and user-friendly hardware, consisting of Arduino Due, extended EMG shields, and MPU-6050, shows promise for practical PSL communication.

A thorough analysis of wearable sensor-based systems for SLR was done by Kudrinko et al. [22]. The analysis covered 72 studies that looked at factors such sensor set-up, recognition model, lexicon size, and identification accuracy between 1991 and 2019. The assessment found problems with sign border detection, scalability to bigger lexicons, and model convergence as well as gaps in the field. The study's findings may help in the creation of wearable sensor-based sign language recognition technology. To make these devices as comfortable for users as possible, wireless transmission, and Deaf users' input must all be taken into account.

A unique multimodal framework for sensor-based SLR combining Microsoft Kinect and Leap motion sensors was proposed by Kumar et al. [23]. Their system extracts elements for recognition by capturing finger and palm postures from different viewpoints. Separate classifiers using the Hidden Markov Model (HMM) and the Bidirectional Long Short-Term Memory Neural Network (BLSTM-NN) were utilised, and their outcomes were integrated to increase accuracy. Testing on a dataset of 7500 ISL gestures showed that fusing data from both sensors outperformed single sensor-based recognition, with accuracies of 97.85% and 94.55% achieved for single and double-handed signs, respectively. The study emphasizes the robustness and potential of the proposed multimodal framework for SLR systems.

Brazilian Sign Language feature extraction using RGB-D sensors was proposed by Moreira Almeida et al. [24]. From RGB-D photos, they derived seven vision-based traits and connected them to structural aspects based on hand movement, shape, and location. Support vector machines (SVM) were used to classify signs, and the average accuracy was above 80%. By characterising each sign language's unique phonological structure, the concept may be applied to other sign language systems and shows potential for SLR systems. RGB-D sensors have the potential to improve image processing algorithms and recognise hand gestures.

Amin et al. [25] conducted a comparative review on the applications of different sensors for SLR. Their review focused on various techniques and sensors used for SLR, with an emphasis on sensor-based smart gloves for capturing hand movements. The authors analyzed existing systems, categorized authors based on their work, and discussed trends and deficiencies in SLR. The comparative analysis provides valuable insights for researchers and offers guidance for developing translation systems for different sign languages. Additionally, the review emphasizes the potential of generated datasets from these sensors for gesture recognition tasks.

In their study, Rajalakshmi et al. [26] proposed a hybrid deep neural net methodology for recognizing Indian and Russian sign gestures. The system used a variety of methods to extract multi-semantic features like non-manual components and manual co-articulations, such as 3D deep neural net with atrous convolutions, attention-based Bi-LSTM, and modified autoencoders. The hDNN-SLR achieved accuracies of 98.75%, 98.02%, 97.94%, and 97.54% for the respective WLASL datasets, surpassing other baseline architectures.

*B. Vision- based approach*

In a study by Matyáš Boháček et al. [1], a transformer-based neural network was employed for word-level SLR. The authors achieved an accuracy of 63.18% on the WLASL dataset and 100% on the LSA64 dataset, focusing on pose-based recognition using transformers.

Another study by Atyaf Hekmat Mohammed Ali et al. [2] developed a real-time SLR system using a Convolutional Neural Network (CNN) with the SqueezeNet module for feature extraction. They achieved 100% accuracy in off-time testing and 97.5% accuracy in real-time using the ASL dataset. This approach emphasized real-time recognition and feature extraction using SqueezeNet.

A forthcoming paper by Rajalakshmi E et al. [3] proposed a hybrid approach combining transformer-based neural networks and CNNs for continuous SLR and translation. Although accuracy results are not yet reported, this approach stands out for its use of a hybrid neural network and translation capabilities, leveraging the ISLW dataset and the Phoenix14T Weather dataset.

In a study by Aashir Hafeez et al. [4], a SLR system was developed using various deep learning techniques, including Artificial Neural Network (ANN), K-Nearest Neighbor (KNN), SVM, and CNN. Accuracy rates of 41.95%, 60.79%, 84.18%, and 88.38% were achieved using the ASL dataset, focusing on improving accuracy rates using multiple deep learning techniques.

Katerina Papadimitriou et al. [5] proposed an SLR system using modulated graph convolutional networks. While no accuracy results were reported, this approach stood out for its utilization of 3D convolutions and modulated graph convolutional networks.

Kaushal Goyal et al. [6] developed a SLR system using LSTM and CNN, achieving accuracy rates of 85% and 97% using the ISL dataset. This approach distinguished itself using LSTM and CNN for SLR.

Soumen Das et al. [8] developed an occlusion-robust SLR system using keyframe + VGG19 + BiLSTM, achieving an accuracy of 94.654% using the ISL Publicly Available dataset. This approach focused on occlusion-robust recognition and utilized keyframes.

C J Sruthi et al. [9] developed a deep learning-based SLR system using CNN, achieving an accuracy of 98.6% with the ISL dataset. This approach stood out for its focus on the Indian sign language and the high accuracy achieved using a CNN-based approach. This system holds potential in enhancing accessibility and communication for the deaf community in India.

## III. SIGN LANGUGAE RECOGNITION TECHNIQUES ANALYSIS

SLR research encompasses various techniques and methodologies to improve accuracy and performance in recognizing sign gestures. One approach is the utilization of Transformer-based Neural Networks, as demonstrated by Matyáš Boháček et al. [1]. Their work involved employing a Transformer-based Neural Network for SLR and achieved an accuracy of 63.18% on the WLASL dataset and a perfect accuracy of 100% on the LSA64 dataset. Similarly, Rajalakshmi E et al. [3] incorporated a Transformer-based Neural Network along with CNN in their SLR framework. While the specific accuracy is not mentioned, they used the Word-level ISL (ISLW) dataset and Phoenix14T Weather dataset for evaluation.

Convolutional Neural Networks (CNN) have proven to be effective in SLR. Atyaf Hekmat Mohammed Ali et al. [2] employed CNN with the SqueezeNet module for feature extraction in their research. Their model achieved an outstanding accuracy of 100% in off-time testing and an impressive accuracy of 97.5% in real-time testing on the ASL dataset. Similarly, C J, Sruthi et al. [9] leveraged CNN and obtained a remarkable accuracy of 98.6% on the ISL dataset. Another study by K. Nimisha et al. [14] involved the use of YOLO, PCA, SVM, ANN, and CNN for SLR, achieving an accuracy of 90% on the ASL dataset. Zhou, H., Zhou et al. [15] proposed a multi-cue framework for SLR and translation, employing CNN as part of their approach, and achieved an accuracy of 95.9%.

Long Short-Term Memory (LSTM) networks have also been applied in SLR research. Kaushal Goyal et al. [6] utilized LSTM and CNN in their framework. The LSTM network achieved an accuracy of 85%, while the CNN achieved an accuracy of 97% on the ISL dataset.

Sensor-based approaches have shown promise in SLR. Khomami et al. [21] developed a wearable system using sEMG and IMU sensors, combined with a KNN classifier, for Persian Sign Language (PSL) recognition. Their research achieved an accuracy of 95.03% (SD: 0.76%) and an improved accuracy of 96.13% (SD: 0.46%) on the PSL dataset. Additionally, Kumar et al. [23] incorporated Microsoft Kinect and Leap motion sensors to capture hand movements and employed an HMM and BLSTM classifier. On the ISL dataset, they obtained 97.85% accuracy for single-handed signs and 94.55% accuracy for double-handed signals.

Other techniques in SLR include hybrid approaches and unique methodologies. Aloysius, N. et al. [10] utilized a hybrid approach combining PCNN and GM, focusing on frame-label alignment techniques. Their specific accuracy is not mentioned. Rajalakshmi et al. [26] proposed a 3D deep neural net with atrous convolutions and Attention-based Bi-LSTM for Indo-Russian Sign Language recognition. Their research yielded impressive accuracies of 98.75%, 98.02%, 97.94%, and 97.54% on the WLASL100, WLASL300, WLASL1000, and WLASL2000 datasets, respectively.

These diverse techniques and methodologies reflect the ongoing efforts in the field of SLR to improve the accuracy and performance of sign gesture recognition systems. Each approach brings unique contributions and advancements, contributing to the overall progress of SLR research.

TABLE I: ANALYSIS OF SIGN LANGUAGE DETECTION TECHNIQUE

| Paper | Techniques/Observations | Year | Accuracy Achieved | Dataset |
|---|---|---|---|---|
| Matyáš Boháček et al. [1] | Transformer-based Neural Network | 2022 | WLASL - 63.18%, LSA64 - 100 % | WLASL, LSA64 dataset |
| Atyaf Hekmat Mohammed Ali et al. [2] | CNN with SqueezeNet module for feature extraction | 2022 | 100% in off-time testing, 97.5 % in real-time | ASL[a] |
| Rajalakshmi E et al. [3] | Transformer-based Neural Network, CNN | 2023 | - | Word-level ISL (ISLW) dataset Phoenix14T Weather dataset |
| Aashir Hafeez et al. [4] | ANN, KNN, SVM, CNN | 2023 | 41.95%, 60.79%, 84.18%, 88.38% | ASL[a] |
| Katerina Papadimitriou et al. [5] | 3D-CNN Model | 2023 | - | AUTSLc, ITI GSLd |
| Kaushal Goyal et al. [6] | LSTM, CNN | 2023 | 85% ,97% | ISL[b] |
| Gero Strobel et al. [7] | TNN | 2023 | - | Video Dataset |
| Soumen das et al. [8] | Keyframe + VGG-19 + BiLSTM | 2023 | 94.65% | ISLb Publicly Available |
| C J, Sruthi et al. [9] | CNN | 2019 | 98.6% | ISL[b] |
| Aloysius, N. et al. [10] | Hybrid PCNN and GM. Frame-label alignment technique of Deep learning is used | 2020 | - | - |
| K. Amrutha et al. [11] | Convex Hull - Feature extraction KNN | 2021 | 65% | ASL[a], ISL[b] |
| R. Cui, H. Liu et al. [12] | CNN followed by temporal Convolutional and pooling layers. | 2019 | 91.93% | RWTH phoenix Weather 2014 database |
| Papastratis, I. et al. [13] | 2d-CNN and weights learned from ImageNet dataset. | 2021 | - | RWTH PhoenixWeather-2014 and the CSL[f] and GSL[d] |
| K. Nimisha et al. [14] | YOLO, PCA, SVM, ANN and CNN | 2021 | 90% | ASL[a] |

| Zhou, H., Zhou et al. [15] | Multi-cue framework for SLR and translation. STMC framework | 2021 | 95.9% | - |
|---|---|---|---|---|
| Shirbhate, Radha S. et al. [16] | KNN, SVM for feature extraction | 2020 | 100% | ISL[b] |
| M. Xie et al. [17] | RNN | 2019 | 99.4% | ISL[b] |
| Selvaraj, P. et al. [18] | ST-GCN | 2022 | 94.7% | ISL[b] |
| Kumar, D. A. et al. [19] | Adaptive Graph Matching Intra-GM to extract signs alone, discarding ME | 2018 | 98.32% | ISL[b] |
| Korban, Matthew et al. [20] | HMM and hybrid KNN-DTW algorithm | 2018 | 92.4% | PSL[e] |
| Khomami et al. [21] | sEMG and IMU sensors, KNN classifier | 2020 | 95.03% (SD: 0.76%), 96.13% (SD:0.46%) | PSL[e] |
| Kudrinko et al. [22] | wearable sensor-based system | 2020 | - | ASL[a], ISL[b] |
| Kumar et al. [23] | Microsoft Kinect and Leap motion sensors to record hand movement, HMM and BLSTM classifier | 2016 | 97.85% for single-hand signs, 94.55% for double-handed signs | ISL[b] |
| Moreira Almeida et al. [24] | RGD-B sensor, SVM classifier | 2014 | 80% | Brazilian Sign-Language |
| Amin et al. [25] | sensor-based smart gloves | 2022 | - | ASL[a] |
| Rajalakshmi et al. [26] | 3D deep neural net with atrous convolutions and Attention-based Bi-LSTM. | 2023 | 98.75% on WLASL100, 98.02% on WLASL300, 97.94% on WLASL1000, 97.54% on WLASL2000 | Indo-Russian Sign Language Dataset |

[a.] American Sign-Language
[b.] Indian Sign-Language
[c.] Ankara University Turkish Sign Language Dataset
[d.] German Sign-Language
[e.] Persian Sign-Language
[f.] Chinese Sign-Language

Based on extensive inquiry and analysis, the authors provide several prospective routes for beginners, academics, and researchers to take in order to further their work. Some areas that need improvement involve:

Tiny sample sizes: Some researchers have trained and evaluated their models using tiny datasets, which may not fully reflect the variety of sign languages and signers.

Cultural prejudice: Several studies have concentrated on identifying sign languages in particular cultural contexts, which may not be relevant to other cultures or nations.

Measurement biases: The use of certain measurement techniques, which may not be completely trustworthy or valid, may have an impact on the findings of some research. Lack of control groups makes it challenging to assess whether the success of the model is primarily attributable to the algorithm or to other variables. Some studies have not included a control group in their analyses.

Studies that are restricted to a few sign languages or simple signs may not be relevant to other sign languages or more complicated signs. Studies that are restricted to a few sign languages or simple signs.

## IV. FUTURE SCOPE

*Accessibility:* SLR models can help the community become more accessible by allowing the deaf population to communicate more effectively with hearing individuals.

*Education:* Sign language recognition models may be used to create instructional materials and software that will make it easier for people to learn sign language.

*Healthcare:* To facilitate communication between patients who use sign language and healthcare professionals, SLR models can be employed in healthcare settings.

*Entertainment:* SLR models may be used to provide more accessible entertainment materials, such as sign language interpretation in motion pictures and television programmes.

*Interaction with Autonomous Systems:* Models for the recognition of sign language can be used to develop robots and other autonomous systems that can interact with people who communicate primarily through sign language.

*Accessibility in public areas:* To make public areas like railway stations, airports, and other transportation hubs more accessible for those who use sign language, sign language recognition models can be utilised.

*Services for sign language interpretation:* By using SLR models, it is possible to provide more precise and effective sign language interpretation services while also lowering the demand for human interpreters.

## V. CONCLUSION

SLR is a significant research area with implications for communication, accessibility, and inclusion for the deaf community. The analysis of recent developments in SLR techniques highlights a range of approaches, including glove-based and vision-based methods.

Glove-based approaches utilize wearable devices to capture hand movements and gestures, while vision-based approaches leverage computer vision algorithms to analyze visual cues. Both approaches have shown promising results in recognizing sign gestures.

In the glove-based approach, wearable sensor-based systems using sensors such as sEMG and IMUs have been developed. These systems can accurately capture hand movements and gestures, leading to high recognition accuracies. However, challenges still exist in terms of sign boundary detection and scalability to larger sign lexicons. Future research in this area should focus on addressing these challenges and ensuring user comfort and acceptance.

Vision-based approaches, particularly those using deep learning techniques, have also achieved significant progress in sign language recognition. CNNs have been widely used for feature extraction and classification of sign gestures, demonstrating high accuracies. Other deep learning techniques like LSTM networks and Transformer-based Neural Networks have also been explored, showing promising results in capturing the sequential nature of sign gestures.

Sensor-based approaches utilizing RGB-D sensors, Microsoft Kinect, or Leap motion sensors have also contributed to sign language recognition. These technologies provide valuable information about hand movements and enable accurate feature extraction. However, they may come with additional hardware requirements and cost considerations.

Hybrid approaches that combine different methodologies, such as deep neural networks, pulse-coupled neural networks (PCNN), graph matching (GM), and frame-label alignment techniques, have also shown improved performance in recognizing sign gestures. These hybrid approaches leverage the strengths of different techniques and provide a more comprehensive solution.

Overall, the field of SLR is advancing rapidly, driven by advancements in machine learning, computer vision, and wearable technologies. Future research should focus on addressing the challenges in sign boundary detection, scalability, user comfort, and acceptance. Additionally, efforts should be made to include the perspectives and input of the deaf community in the development of SLR systems to ensure their effectiveness and suitability for real-world applications.

By developing accurate and robust SLR systems, we can enhance communication accessibility for the deaf and community, promote inclusivity, and facilitate their participation in various domains of life.

## VI. ACKNOWLEDGMENT

As we conclude this research, we gratefully acknowledge the invaluable support and contributions of numerous individuals and organizations. We extend our sincere appreciation to all those who have been part of this journey.

We wish to express our deepest gratitude to Dr. S.K Singh, our esteemed mentor, for his unwavering guidance, unwavering support, and constant motivation throughout this endeavour. His wealth of expertise and experience were instrumental in helping us achieve our objectives and surmount any obstacles that arose. Dr. S.K Singh has been an invaluable member of our team, and we are profoundly grateful for his outstanding contributions.

## REFRENCES

[1] M. Boháček and M. Hrúz, "Sign Pose-based Transformer for Word-level Sign Language Recognition," in 2022 IEEE/CVF Winter Conference on Applications of Computer Vision Workshops (WACVW), 4-8 January 2022. DOI: 10.1109/WACVW54805.2022.00024.

[2] Mohammedali, A. H., Abbas, H. H., & Shahadi, H. I. (2022). Real-time sign language recognition system. International Journal of Health Sciences, 6(S4), 10384–10407. https://doi.org/10.53730/ijhs.v6nS4.12206.

[3] E. Rajalakshmi, R. Elakkiya, V. Subramaniyaswamy, K. Kotecha, M. Mehta, and V. Palade, "Continuous Sign Language Recognition and Translation Using Hybrid Transformer-Based Neural Network," available at SSRN: https://ssrn.com/abstract=4424708 or http://dx.doi.org/10.2139/ssrn.4424708.

[4] A. Hafeez, S. Singh, U. Singh, P. Agarwal, and A. K. Jayswal, "Sign Language Recognition System Using Deep-Learning for Deaf and Dumb," International Research Journal of Modernization in Engineering Technology and Science, vol. 5, no. 4, pp., Apr. 2023. DOI: 10.56726/IRJMETS36063.

[5] K. Papadimitriou and G. Potamianos, "Sign Language Recognition via Deformable 3D Convolutions and Modulated Graph Convolutional Networks," in Proceedings of the IEEE International Conference on Acoustics, Speech, and Signal Processing (ICASSP), 2023.

[6] Dr. V. G and K. Goyal, "Indian Sign Language Recognition Using Mediapipe Holistic." arXiv, 2023. doi: 10.48550/ARXIV.2304.10256.

[7] Strobel, G., Schoormann, T., Banh, L., & Möller, F. (in press). Artificial Intelligence for Sign Language Translation – A Design Science Research Study. Communications of the Association for Information Systems, 52, pp-pp. Retrieved from https://aisel.aisnet.org/cais/vol52/iss1/33

[8] S. DAS, S. kr. Biswas, and B. Purkayastha, "Occlusion Robust Sign Language Recognition System for Indian Sign Language Using CNN and Pose Features." Research Square Platform LLC, Apr. 20, 2023. doi: 10.21203/rs.3.rs-2801772/v1.

[9] S. C J and L. A, "Signet: A Deep Learning based Indian Sign Language Recognition System," 2019 International Conference on Communication and Signal Processing (ICCSP). IEEE, Apr. 2019. doi: 10.1109/iccsp.2019.8698006.

[10] Aloysius, N., Geetha, M. Understanding vision-based continuous sign language recognition. Multimed Tools Appl 79, 22177–22209 (2020). https://doi.org/10.1007/s11042-020-08961-z

[11] K. Amrutha and P. Prabu, "ML Based Sign Language Recognition System," 2021 International Conference on Innovative Trends in Information Technology (ICITIIT), 2021, pp. 1-6.

[12] R. Cui, H. Liu and C. Zhang, "A Deep Neural Framework for Continuous Sign Language Recognition by Iterative Training," in IEEE Transactions on Multimedia, vol. 21, no. 7, pp. 1880-1891, July 2019.

[13] Papastratis, I.; Dimitropoulos, K.; Daras, P. Continuous Sign Language Recognition through a Context-Aware Generative Adversarial Network. Sensors 2021, 21, 2437.

[14] K. Nimisha and A. Jacob, "A Brief Review of the Recent Trends in Sign Language Recognition," 2020 International Conference on Communication and Signal Processing (ICCSP), 2020, pp. 186-190.

[15] Zhou, H., Zhou, W., Zhou, Y., & Li, H. (2021). Spatial-Temporal Multi-Cue Network for Sign Language Recognition and Translation. IEEE Transactions on Multimedia.

[16] Shirbhate, Radha S., Mr. Vedant D. Shinde, Ms. Sanam A. Metkari, Ms. Pooja U. Borkar and Ms. Mayuri A. Khandge. "Sign language Recognition Using Machine Learning Algorithm." (2020).

[17] M. Xie and X. Ma, "End-to-End Residual Neural Network with Data Augmentation for Sign Language Recognition," 2019 IEEE 4th Advanced Information Technology, Electronic and Automation Control Conference (IAEAC), 2019, pp. 1629-1633.

[18] Selvaraj, P., Gokul N., C., Kumar, P., & Khapra, M.M. (2022). OpenHands: Making Sign Language Recognition Accessible with Pose-based Pretrained Models across Languages. ArXiv, abs/2110.05877.

[19] Kumar, D. A., Sastry, A. S. C. S., Kishore, P. V. V., & Kumar, E. K. (2018). Indian sign language recognition using graph matching on 3D motion captured signs. Multimedia Tools and Applications.

[20] Korban, Matthew & Nahvi, Manoochehr. (2018). An algorithm on sign words extraction and recognition of continuous Persian sign language based on motion and shape features of hands. Formal Pattern Analysis & Applications. 21. 10.1007/s10044-016-0579-2.

[21] Khomami, S. A., & Shamekhi, S. (2021). Persian sign language recognition using IMU and surface EMG sensors. In Measurement

(Vol. 168, p. 108471). Elsevier BV. https://doi.org/10.1016/j.measurement.2020.108471

[22] K. Kudrinko, E. Flavin, X. Zhu, and Q. Li, "Wearable Sensor-Based Sign Language Recognition: A Comprehensive Review," IEEE Reviews in Biomedical Engineering, vol. 14. Institute of Electrical and Electronics Engineers (IEEE), pp. 82–97, 2021. doi: 10.1109/rbme.2020.3019769.

[23] P. Kumar, H. Gauba, P. Pratim Roy, and D. Prosad Dogra, "A multimodal framework for sensor based sign language recognition," Neurocomputing, vol. 259. Elsevier BV, pp. 21–38, Oct. 2017. doi: 10.1016/j.neucom.2016.08.132

[24] S. G. Moreira Almeida, F. G. Guimarães, and J. Arturo Ramírez, "Feature extraction in Brazilian Sign Language Recognition based on phonological structure and using RGB-D sensors," Expert Systems with Applications, vol. 41, no. 16. Elsevier BV, pp. 7259–7271, Nov. 2014. doi: 10.1016/j.eswa.2014.05.024

[25] M. S. Amin, S. T. H. Rizvi, and Md. M. Hossain, "A Comparative Review on Applications of Different Sensors for Sign Language Recognition," Journal of Imaging, vol. 8, no. 4. MDPI AG, p. 98, Apr. 02, 2022. doi: 10.3390/jimaging8040098.

[26] E. Rajalakshmi et al., "Multi-Semantic Discriminative Feature Learning for Sign Gesture Recognition Using Hybrid Deep Neural Architecture," IEEE Access, vol. 11. Institute of Electrical and Electronics Engineers (IEEE), pp. 2226–2238, 2023. doi: 10.1109/access.2022.3233671.